\documentclass[a4paper]{llncs}

\usepackage[utf8]{inputenc}
\usepackage[english]{babel}
\usepackage{amsmath}
\usepackage{amsfonts}
\usepackage{float}
\usepackage{subfig}
\usepackage{graphicx}
\usepackage{authblk}
\usepackage{color}

\usepackage[T1]{fontenc}
\usepackage[utf8]{inputenc}

\graphicspath{{figure/}}

\title{Computationally efficient cardiac views projection using 3D Convolutional Neural Networks}

\author{
	Matthieu Le\inst{1} \and Jesse Lieman-Sifry\inst{1} \and 
	Felix Lau\inst{1} \and Sean Sall\inst{1} \and 
	Albert Hsiao\inst{1} \and Daniel Golden\inst{1}
}

\institute{Arterys Inc.}

\date{} 

\begin{document}

\maketitle

\begin{abstract} 
4D Flow is an MRI sequence which allows acquisition of 3D images of the heart. The data is typically acquired volumetrically, so it must be reformatted to generate cardiac long axis and short axis views for diagnostic interpretation. These views may be generated by placing 6 landmarks: the left and right ventricle apex, and the aortic, mitral, pulmonary, and tricuspid valves. In this paper, we propose an automatic method to localize landmarks in order to compute the cardiac views. Our approach consists of first calculating a bounding box that tightly crops the heart, followed by a landmark localization step within this bounded region. Both steps are based on a 3D extension of the recently introduced ENet. We demonstrate that the long and short axis projections computed with our automated method are of equivalent quality to projections created with landmarks placed by an experienced cardiac radiologist, based on a blinded test administered to a different cardiac radiologist.  
\end{abstract}

\section{Introduction}

\begin{figure}
  \centering
  \includegraphics[width=0.65\linewidth]{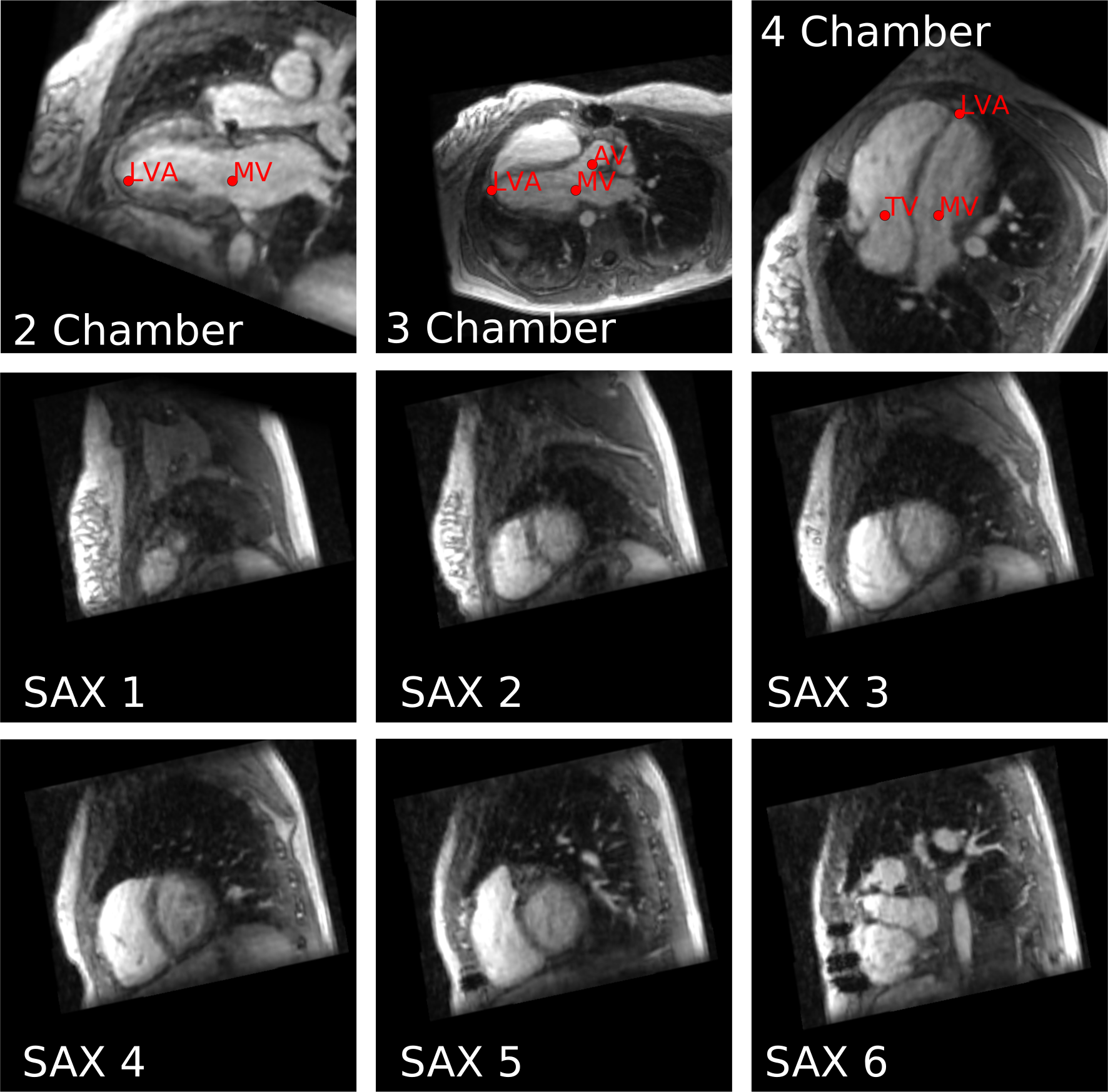}
  \caption{\small{First row: 2 chamber, 3 chamber, and 4 chamber views; second and third rows, left to right, then top to bottom: 6 slices of the SAX views from the apex to the base for the left ventricle.}}
  \label{fig:cardiac_views}
\end{figure}

Cardiac pathologies are often best evaluated along the principle axes of the heart, on long and short axis views. Because 4D Flow scans are typically acquired volumetrically, these planar views must be reconstructed for interpretation by a cardiac radiologist. In order to generate the standard cardiac views, the user may manually place 6 cardiac landmarks: the left and right ventricle apex (LVA, and RVA), and the aortic, mitral, pulmonary, and tricuspid valves (AV, MV, PV, and TV). With training, locating the landmarks using an interactive clinical tool takes a trained radiologist roughly 3 minutes. The landmark positions are then used to compute the 2, 3, and 4 chamber views, as well as the short axis stack (Fig. \ref{fig:cardiac_views}). Although these views can be computed for the left and right ventricles, for the sake of conciseness, we focus on the left ventricle cardiac views. The 4 chamber view is defined as the plane going through the TV, MV, and LVA, with the LVA oriented on the top of the image. The 3 chamber view is defined as the plane going through the AV, MV, and LVA, with the LVA placed on the left of the image. The 2 chamber view is defined as the plane bisecting the large angle of the 3 and 4 chamber planes, with the LVA placed on the left of the image. The SAX (short axis stack views) are generated by sampling planes orthogonal to the axes defined by the LVA and MV, with the left ventricle in the center of the image, and the right ventricle on the left of the image. 

Methods that utilize deep learning techniques to determine keypoints in 2D or 3D images have recently shown state of the art results. Payer et al. \cite{payer2016regressing} proposed a method to detect landmarks in 2D and 3D medical images by regressing heat maps of the landmark location. Other approaches have been suggested for landmark detection such as using reinforcement learning \cite{ghesu2016artificial}, where an agent travels through the image and stops when a landmark is found. Some approaches proposed to detect cardiac landmarks starting with left ventricle segmentations \cite{lu2011automatic}.

In this paper, we build on \cite{pfister2015flowing,payer2016regressing} and introduce a method to automatically compute cardiac views based on a landmark detection network. Our approach may obviate manual placement of landmarks, to fully automate computation of cardiac views. We demonstrate that the long and short axis projections computed with our automated method are of equivalent quality to those created by expert annotators, based on a blinded test administered to a board-certified cardiac radiologist. We finally show that our model can be applied to other cardiac MRI sequences, such as volumetrically-acquired 3D-cine single breath hold MRI.

\section{Methods}

\begin{figure}
  \centering
  \includegraphics[width=\linewidth]{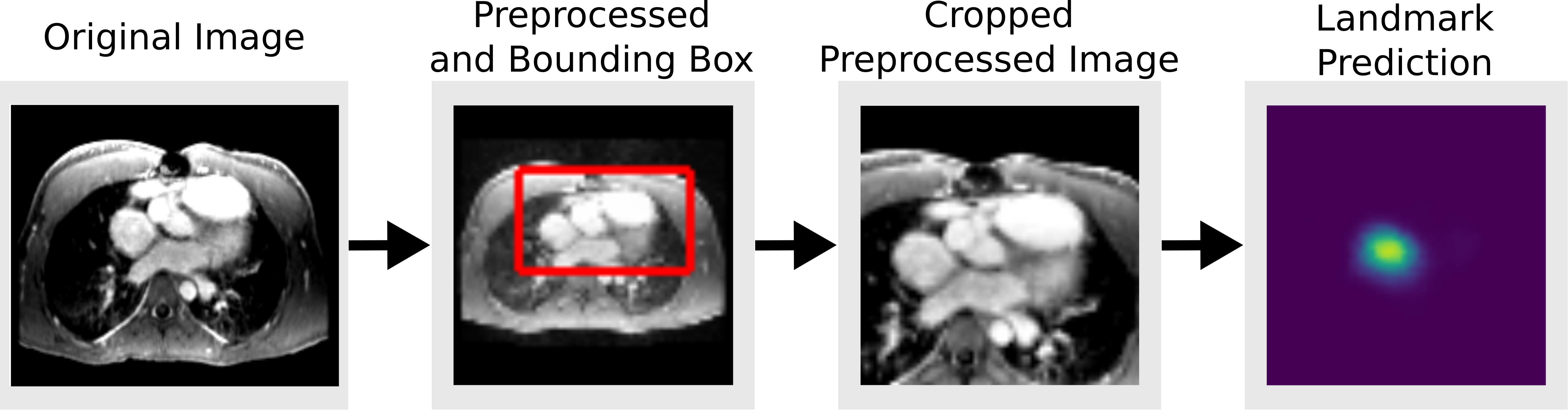}
  \caption{\small{From left to right: original input, preprocessed image for bounding box prediction, cropped preprocessed input, landmark detection (here the aortic valve).}}
  \label{fig:pipeline}
\end{figure}

\begin{figure}
  \centering
  \includegraphics[width=0.9\linewidth]{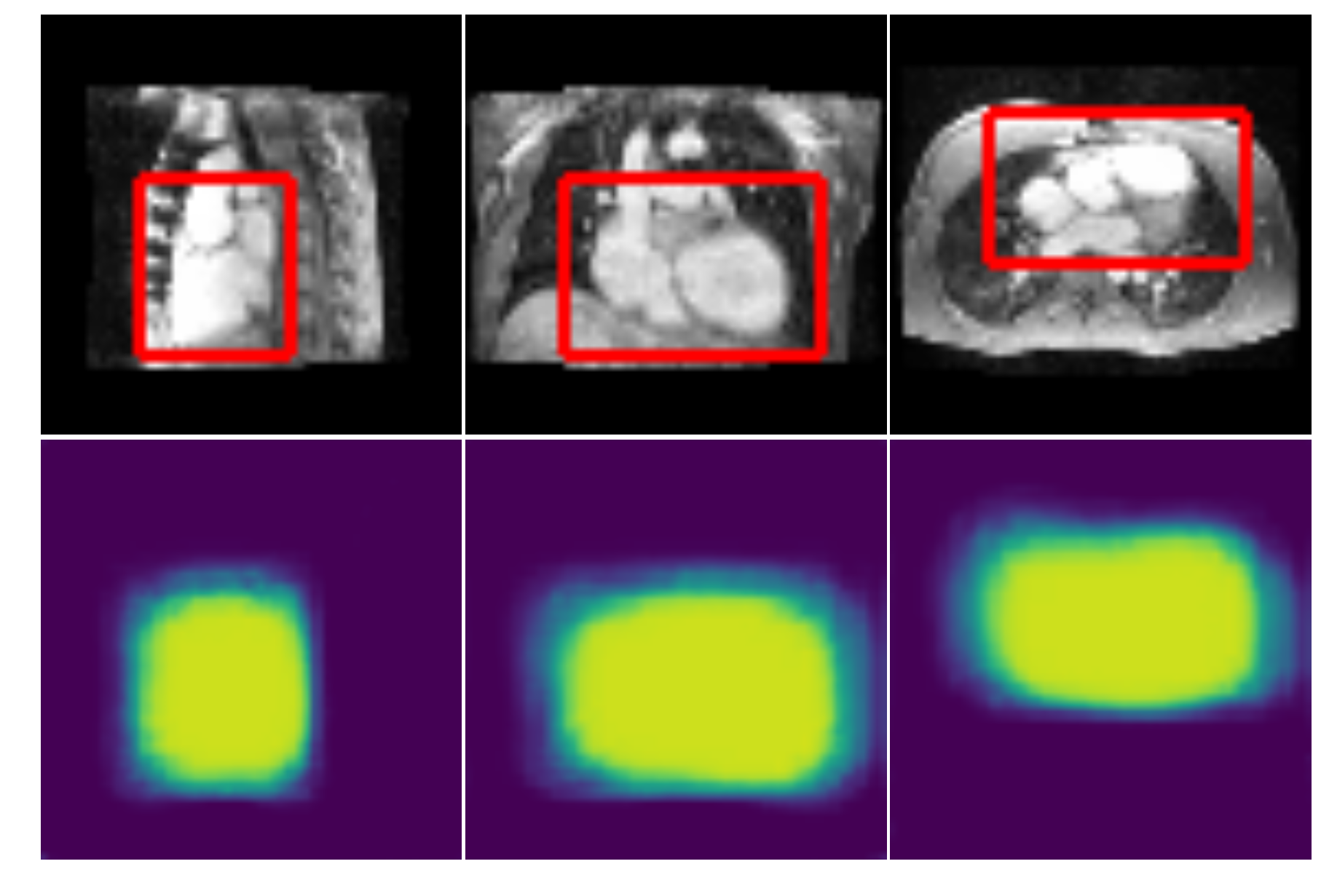}
  \caption{\small{Bounding box prediction examples for one 3D image of the test set. The anatomy image is on top, the output of the network on the bottom. The red lines outline the inferred bounding box from the output map. From left to right, sagittal, coronal, and axial views.}}
   \label{fig:results_bounding_box}
\end{figure}

\textbf{Pipeline.} We utilize a two-step approach to locate the six landmarks that define the cardiac views. First, we predict a tight bounding box around the heart. Second, we use the cropped image to predict the landmark location (Fig. \ref{fig:pipeline}). For both steps, we use a 3D ENet architecture based on the original 2D implementation from \cite{paszke2016enet}. For bounding box prediction, the input is the 3D image, and the output is a semantic segmentation map of the volume within the bounding box (Fig. \ref{fig:results_bounding_box}). We convert the bounding box segmentation map to bounding box coordinates by taking, per axis, the location of the 5th and 95th percentiles of the cumulative sum over the non-considered axis. For landmark prediction, the input of the network is a crop around the heart of the original 3D image and the output is a 6 channel heat map (one channel per landmark), where the location of the peak corresponds to the predicted landmark position (Fig. \ref{fig:results_landmarks}).

\textbf{Data.} For bounding box detection, the database comprises 81 unique patients. For each patient, an average of 2.9 time points were manually annotated with a bounding box enclosing the heart, resulting in 234 distinct 3D images. The ground truth for the network is a binary mask equal to one within the manually defined bounding boxes, and zero elsewhere. For landmark detection, the database comprises 310 unique patients. For each patient, an average of 2.1 time points were annotated by expert cardiac radiologists, resulting in 664 distinct 3D images. Note that for each image, not all six landmarks are necessarily annotated (Table \ref{table:landmarks_results}). The ground truth is a 6-channel 3D heat map where the location of each landmark is encoded by a 3D isotropic Gaussian with a fixed variance of 4 voxels.

\textbf{Preprocessing.} Before both bounding box and landmark detection, we apply the following preprocessing: i) resize the image to an isotropically re-sampled $64 \times 64 \times 64$ cube, padding with zero if necessary, ii) clip the image pixel intensities between the 1st and 99th percentile intensities, iii) normalize the image intensity to be between 0 and 1, iv) apply adaptive histogram \cite{zuiderveld1994contrast}, v) center and scale the image. Figure \ref{fig:pipeline} shows the result of the pre-processing for the bounding box and landmark detection steps.

\textbf{3D ENet.} In both steps we use a 3D extension of the 2D ENet neural network \cite{paszke2016enet}. It is a fully convolutional network composed of a downsampling path and an upsampling path that produces segmentation maps at the same resolution as the input image. ENet makes use of recent advances in deep learning for improved computational efficiency  \cite{paszke2016enet}. The network architecture utilizes early downsampling and a smaller expanding path than contracting path. It also makes use of bottleneck modules, which are convolutions with a small receptive field that are applied in order to project the feature maps into a lower dimensional space in which larger kernels can be applied \cite{he2016deep}. Finally, throughout the network, ENet leverages a diversity of low cost convolution operations such as asymmetric ($1 \times 1 \times n$, $1 \times n \times 1$, and $n \times 1 \times 1$) convolutions and dilated convolutions \cite{yu2015multi}.

\textbf{Training.} We split the data such that 80\% of patients are in the training set, 10\% in the validation set, and 10\% in the test set, such that a patient images are only present in one set. For the bounding box ENet, we use the parameters described in \cite{paszke2016enet}. We train it using the Adam optimizer \cite{kingma2014adam} with pixelwise cross-entropy loss. 
We train the landmark detection models with the Adam optimizer and an l2 loss. We only compute the loss corresponding to the landmarks present in the considered annotations. This allows to train for all 6 landmarks using every 3D image even though not all images have ground truth for all 6 landmarks.
We run a hyperparameter search in which we train 50 models for 40 epochs and pick the 3 models with the lowest l2 loss. We then re-train the three best models for 100 epochs. We then designate the best model as the one with the lowest median error over all landmarks. We search over: the level of the applied distortions (whether or not to flip the images, the intensity of the Gaussian noise, the elastic distortions, affine transformations, brightness, contrast, and blur parameters), the size of the asymmetric convolutions, the number of 1.x bottlenecks, the number of repeated section 2, the number of initial filters, the type of pooling (max or average), the projection scale at which the dimensionality is reduced in a bottleneck, the learning rate, and the amount of dropout (see \cite{paszke2016enet} for details on the naming conventions). We additionally tested the use of skip connections and pyramidal scene parsing module \cite{zhao2016pyramid}, but did not find it improved the results.

\section{Results}

\begin{figure}
  \centering
  \includegraphics[width=\linewidth]{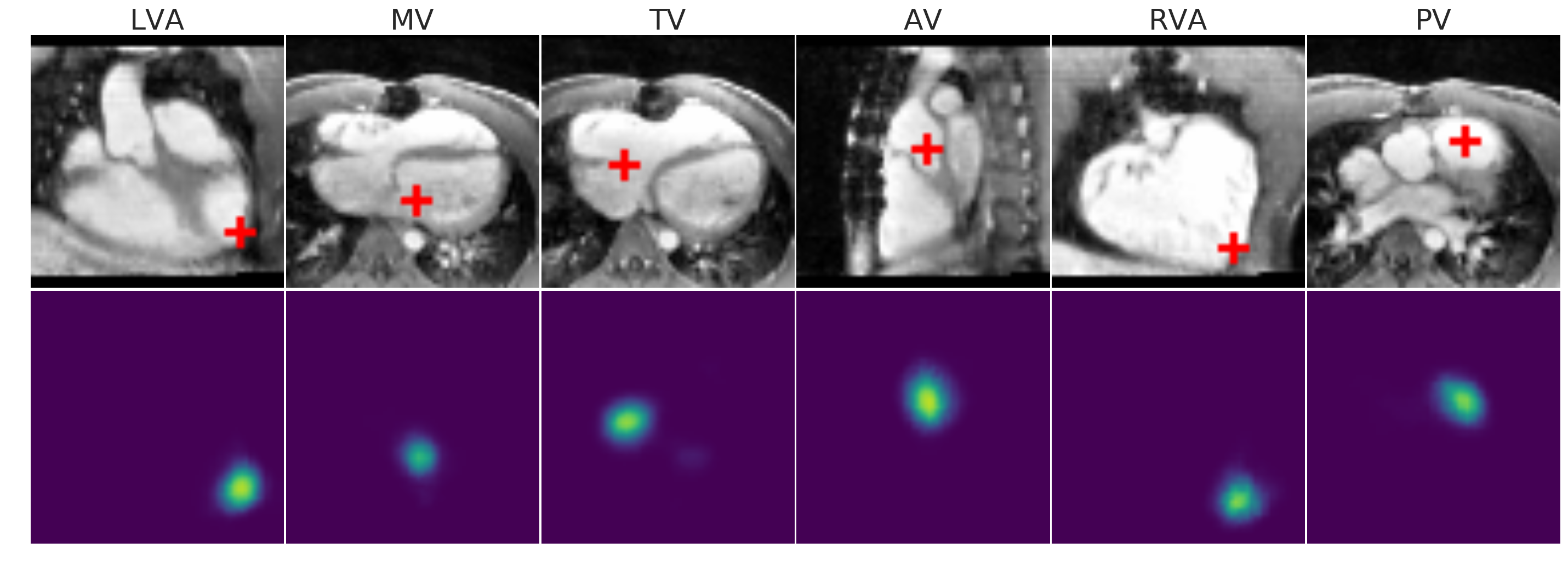}
  \caption{\small{Example of prediction of the 6 landmarks for one 3D image of the test set. The anatomy image is on top, the output of the network on the bottom. For each landmark, we show one 2D slice of the 3D input and output. The red cross on the anatomy image localizes the peak of the output map of the network for the different landmarks. From left to right, the views are coronal, axial, axial, sagittal, coronal, and axial.}}
  \label{fig:results_landmarks}
\end{figure}

\textbf{Bounding Box.} The goal of the bounding box is to be able to crop a region around the heart containing the landmarks. This allows to feed higher resolution images to the subsequent landmark detection network. For this reason, and because the exact extent of a ``good'' bounding box is subject to interpretation, the classical measures of accuracy for semantic segmentation problems are not well suited. Rather we are interested in making sure that the landmarks are within the bounding boxes, and that they are cropping the original image by a sufficient amount. On the landmark database, the bounding boxes are on average $18\%$ the volume of the original image, and $98\%$ of all ground truth landmarks are enclosed within the bounding boxes. The bounding box validation per pixel accuracy is 95\%, and the validation Dice coefficient is 0.83.

Figure \ref{fig:results_bounding_box} shows the results of the bounding box prediction for one patient, and how we compute bounding box coordinates (outlined in red on the top row) from the output segmentation map (bottom row). Note that because the image is then resized to a cube with isotropic spacing for landmark prediction, the final cropped image is often larger than the predicted bounding box.

\textbf{Landmarks.} The results of the designated best model on the test set are presented in Table \ref{table:landmarks_results}. The per landmark median error (distance between the ground truth and the prediction), averaged over all 6 landmarks, is 8.9 mm, and the median error over all prediction is 8.8 mm for the test set. Cadaveric studies by \cite{ilankathir1cadaveric} find that the  average radius of the smallest valve (the PV) is 10.8 mm. Our average localization error is therefore less than the average radius of the smallest cardiac valve.
Figure \ref{fig:results_landmarks} presents the landmark detection results for one patient. The inference time for one study is on average a little less than a second on an NVIDIA Maxwell Titan X, depending on the original size of the input image. 

\begin{table}
  \centering
  \begin{tabular}{|c|c|c|c|}
    \hline
       & Train (nbr of annot.) & Validation (nbr of annot.) & Test (nbr of annot.) \\
    \hline
      LVA & 6.6 (403) & 5.5 (50) & 7.1 (62) \\
    \hline
      MV & 8.4 (514) & 8.8 (59) & 9.5 (68) \\
    \hline
      AV & 6.3 (284) & 6.2 (35) & 6.5 (34) \\
    \hline
      RVA & 7.5 (453) & 6.0 (48) & 9.3 (45) \\
    \hline
      TV & 8.9 (513) & 10.5 (64) & 10.6 (67) \\
    \hline
      PV & 8.8 (326) & 13.6 (34) & 10.2 (34) \\
    \hline
      Average Median Error & 7.7 & 8.4 & 8.9 \\
    \hline
      Median Error & 7.7 & 7.9 & 8.8 \\
    \hline
  \end{tabular}
  \caption{\small{Result of landmarks detection error, and number of annotations in the database. The error is defined as the distance between the ground truth and the prediction in mm.}}
  \label{table:landmarks_results}
\end{table}

\textbf{Cardiac Views.} Figure \ref{fig:cardiac_views} shows the predicted cardiac views for a single time point for the left ventricle. Although the model is trained at sparsely annotated time points of the cardiac cycle (mainly the time points corresponding to the end diastole and the end systole), we can predict the landmarks location for each time point. 

In order to compute the cardiac views, we predict each landmark position at each time point, and then take the median position over time in order to compute the projection with fixed landmark (otherwise, the projected cardiac views would move during the cycle, which can be confusing for the clinician). To evaluate the quality of the cardiac views, we blindly presented them to a board-certified cardiac radiologist who was not involved in the ground truth collection or any other aspect of this work. We presented the radiologist with 2D cine movies over the cardiac cycle of projections computed from both the ground truth annotations and the predicted landmarks for the full test set. The order of the movies were randomized so the clinician did not know whether any given movie came from ground truth or predicted landmarks. We asked the clinician to grade each projections (2/3/4 chamber and SAX views) from 1 (bad projection unsuitable for diagnosis) to 5 (clinically relevant cardiac views). Table \ref{table:cardiac_views} shows the results. The grade shows that the quality of the predicted cardiac views is on par with the ground truth views with an average grade difference of 0.5 between the two, in favor of the predicted cardiac views. The failure mode of the cardiac view computations mainly include the presence of part of the right ventricle in the left ventricle 2 chamber views. 

\begin{figure}
  \centering
  \includegraphics[width=0.8\linewidth]{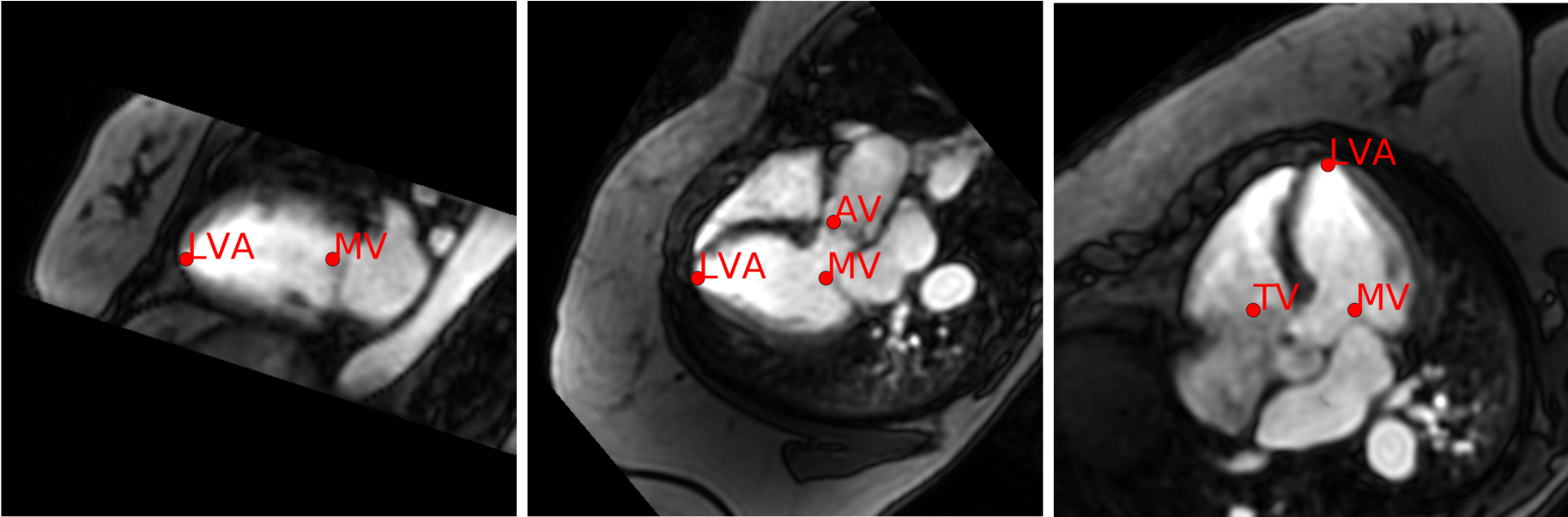}
  \caption{\small{Cardiac views results for a 3D-cine single breath hold MRI. Left: 4 chamber view where the prediction is successful. Middle: 3 chamber view at a time point where the prediction is successful. Right: 3 chamber view at a time point where the prediction is failing.}}

  \label{fig:cardiac_views_cine}
\end{figure}

\textbf{Other MR Sequences.} We show on Figure \ref{fig:cardiac_views_cine} the results of the cardiac views computation using a 3D-cine single breath hold MRI, which is a different sequence than 4D flow. We can see that the learned model generalizes well to similar MRI modalities. 

\begin{table}
  \centering
  \begin{tabular}{|c|c|c|c|}
    \hline
       & 1 - Automated Views & 2 - Radiologist Views & Difference 1 - 2\\
    \hline
      2 Chamber & 4.0 & 3.6 & 0.4\\
    \hline
      3 Chamber & 4.7 & 4.6 & 0.1\\
    \hline
      4 Chamber & 4.8  & 4.1 & 0.7\\
    \hline
      SAX & 4.1  & 4.1 & 0.0\\
    \hline
      Average & 4.4  & 4.1 & 0.5\\
    \hline
  \end{tabular}
  \caption{\small{Grade of the cardiac view evaluation by a board-certified cardiac radiologist.}}
  \label{table:cardiac_views}
\end{table}

\section{Conclusion}

We presented a method to automatically compute long and short axis views for volumetrically acquired cardiac MRI. We showed that our error approaches anatomical variability, and we confirmed the efficacy of our approach with quality ratings in a blinded comparison test from a board-certified cardiac radiologist. The presented approach may obviate manual placement of landmarks, to fully automate computation of cardiac views.
Future work could explore the use of non-cubic cropped images fed to the landmark detection network since it is fully convolutional. We will also investigate the incorporation of flow information to help localize the landmarks. This additional data would likely be most helpful to locate the valves.  However, flow data requires careful handling because it can be particularly noisy.

\bibliographystyle{splncs}
\bibliography{biblio.bib}

\end{document}